\newif\ifabstract
\abstracttrue
\abstractfalse 
\newif\iffull
\ifabstract \fullfalse \else \fulltrue \fi

\documentclass[11pt]{article}
\usepackage[T1]{fontenc}
\usepackage[utf8]{inputenc}
\usepackage{lineno}
\usepackage[table]{xcolor}
\usepackage[colorlinks=false, pdfborder={0 0 0}]{hyperref}
\usepackage{geometry}
\usepackage{natbib}
\usepackage{booktabs}
\usepackage[official]{eurosym}
\usepackage{subcaption}

\usepackage{graphicx}
\usepackage{amssymb}
\usepackage{amsmath,amsfonts,amsthm,bm}
\begin{document}

\title{Discriminative conditional restricted Boltzmann machine for discrete choice and latent variable modelling \footnote{Paper presented at International Choice Modelling Conference 2017}}
\author{Melvin Wong \thanks{Laboratory of Innovations in Transportation (LITrans), Department of Civil Engineering, Ryerson University, Toronto, Canada, Email: {melvin.wong@ryerson.ca}}
	\and Bilal Farooq\thanks{Laboratory of Innovations in Transportation (LITrans), Department of Civil Engineering, Ryerson University, Toronto, Canada, Email: {bilal.farooq@ryerson.ca}}
	\and Guillaume-Alexandre Bilodeau\thanks{Laboratoire d'Interprétation et de Traitement d'Images et Vidéo (LITIV), Department of Computer and Software Engineering, Polytechnique Montr\'eal, Montr\'eal, Canada, Email: {guillaume-alexandre.bilodeau@polymtl.ca}}}

\begin{titlepage}
\maketitle

\thispagestyle{empty}

\begin{abstract}
Conventional methods of estimating latent behaviour generally use attitudinal questions which are subjective and these survey questions may not always be available. We hypothesize that an alternative approach can be used for latent variable estimation through an undirected graphical models. For instance, non-parametric artificial neural networks. In this study, we explore the use of generative non-parametric modelling methods to estimate latent variables from prior choice distribution without the conventional use of measurement indicators. A restricted Boltzmann machine is used to represent latent behaviour factors by analyzing the relationship information between the observed choices and explanatory variables. The algorithm is adapted for latent behaviour analysis in discrete choice scenario and we use a graphical approach to evaluate and understand the semantic meaning from estimated parameter vector values. We illustrate our methodology on a financial instrument choice dataset and perform statistical analysis on parameter sensitivity and stability. Our findings show that through non-parametric statistical tests, we can extract useful latent information on the behaviour of latent constructs through machine learning methods and present strong and significant influence on the choice process. Furthermore, our modelling framework shows robustness in input variability through sampling and validation.
\end{abstract}

\end{titlepage}

\section{Introduction}
\label{S:1}
Complex theories of decision-making processes provide the basis of latent behaviour representation in statistical models focusing on the use of psychometric data such as choice perception and attitudinal questions. Although they can provide important insights into choice processes and underlying heterogeneity, studies have shown the limited flexibility and benefits of statistical latent behaviour models, i.e. Integrated Choice and Latent Variable (ICLV) models \citep{chorus2014possibility,vij2016and}. Two disadvantages are known in ICLV models: first, datasets are required to have attitudinal responses, for instance, likert scale questions in product choice surveys. Second, model mis-specification may occur when latent variable model equations are poorly defined and attitudinal questions are subjective and would change over time.

The objective of this study is to use of machine learning (ML) methods to analyze the underlying latent behaviour in choice models based on a set of synthetic ML considerations and hyperparameters without explicitly using attitudinal or perception attributes. A growing body of behavioural research focuses on patterns and clusters of behaviour characteristics including latent attitudes and choice perceptions. Yet, comparing with specific advanced choice modelling strategies such as ICLV models, our knowledge of the prevalence and consequences of latent behaviour in choice model still remains limited \citep{vij2016and}. Studies of hidden representations using neural network models may give us more nuanced and potentially new perspectives of latent variables on discrete choice experiments and choice behaviour theory \citep{Rungie2012145}. Given the many possible latent variable combinations, it is necessary to use advanced ML techniques to segment population into groups with similar attitudinal profiles. For this study, we have chosen to use restricted Boltzmann machines (RBM). RBM is a non-parametric generative modelling approach that seeks to find latent representations within a homogeneous group by hypothesizing that posterior outputs can be explained with a reduced number of hidden units \citep{le2008representational}. In addition, identifying common latent representation may enable policy makers to better understand the sensitivity and stability of latent behaviour models in surveyed and revealed preference data. We decouple the latent behaviour model underlying the data distribution by estimation on a financial instrument choice behaviour dataset without the need for subjective measurement indicators. The proposed method does not predefine a semantic meaning for each latent variable. Instead, we define a restricted Boltzmann machine to learn the latent relationships and approximate the posterior probability.

We show in our findings that a RBM modelling approach is able to characterize latent variables with semantic meaning without additional psychometric data. The parameters estimated through our RBM model presents strong and significant influence in the choice process. Furthermore, sensitivity analysis have shown that this approach is robust to input data variance and use of generated latent variables improves sampling stability.

The remainder of the paper is organized as follows: in Section \ref{S:2}, we provide a background literature review on latent behaviour models. Section \ref{S:3} describes the conditional RBM modelling approach and model training methodology, given only observed variables without attitudinal questions. Section \ref{S:4} explains the data and the experiment procedure. Section \ref{S:5} presents the results and performance tests. Section \ref{S:6} analyzes the model sensitivity and stability. Finally, section \ref{S:7} discuss the conclusions and future research directions. 

\section{Background}
\label{S:2}
Current practice in choice modelling is targeted at drawing conclusion on the mechanism of the stochastic model and not so much about the nature of the data itself. This leads to simple assumptions of data relevance and statistical properties of explanatory variables \citep{burnham2003model}. A number of parametric and non-parametric modelling methods are available. Parametric models are regression based and random utility maximization structural models. Examples of non-parametric methods include latent class and variable models. Other statistical models include k-means or hierarchical clustering. These non-parametric methods are often criticized for being too descriptive, theoretical, may result in inconsistent estimates and often not possible to make generalizations \citep{ben1999discrete,atasoy2013attitudes,bhat2014new}. Analyzing data through the statistical properties is generally applied for extracting information about the evolution of the responses associated with stochastic input variables rather than having good prediction capabilities. On the other hand, algorithmic modelling approaches such as artificial neural networks (ANN), decision trees, clustering and factor analysis are based on the ability to predict future responses accurately given future input variables within a ‘black-box’ framework \citep{breiman2001statistical}. Econometric choice models can be estimated by using both parametric and non-parametric methods that incorporate machine learning algorithms into discrete choice analysis to learn mappings from latent variables to posterior distribution \citep{eric2008active}.

A number of different approaches which implements the use of attitudinal variables have been used in existing literature \citep{ashok2002extending,morey2006using,hackbarth2013consumer}. The first approach relies on a top-down modelling framework which makes prior assumptions that individuals are divided into multiple market segments and each segment has its own utility function of underlying attributes. In the most generic form, these assumptions are based on multiple sources of unobserved heterogeneity influencing decisions, e.g. inter- and intra-class variance and `agent effect' \citep{yazdizadeh2017generic}.
Fig. \ref{icmc_fig00} illustrates the Latent Class and ICLV model framework which shows the process of deriving latent classes or variables and how it integrates into the structural choice model.

The Latent Class model (LCM) is one such form which assumes a discrete distribution among market segments \citep{hess2014handbook}. LCM derive clusters using a probabilistic model that describes the distribution of the data. Based on this assumption, similarities within a heterogeneous population are identified through assignment of latent class probabilities. Individuals in the same class share a common joint probability distribution among the observed variables. Under assumption of class independence, the utility is generated with a prior hypothesis from several sub-populations, and each sub-population is modelled separately. The resulting classes are often meaningful and easily interpretable. The unobserved heterogeneity in the population is captured by the latent classes, each of which is associated with different utility vector in the sub-model (Fig. \ref{icmc_fig00}a). Another similar class of top-down models are finite mixture models, e.g. Mixed Logit, which allows the parameters to vary with a variance component and that behaviour is dependent on the observable attributes and on the latent heterogeneity which varies with the unobserved factors \citep{hensher2003mixed}.

The use of attitudes and perception latent variables are also particularly interesting and popular in past work \citep{glerum2014forecasting,atasoy2013attitudes}. Choice models with measurement indicator functions treat correlated indicators into multiple latent variables. This factor analysis method is similar to principal component analysis where the latent variables are used as principal components \citep{glerum2014forecasting}. This approach involves the analysis of relationship between indicators and the choice model. Within this domain, there is the sequential and simultaneous estimation process. Sequential approach first estimates a measurement model which derives the relationship between latent variables and indicators. Then, a choice model is estimated, integrating over the distribution of the latent variables. The main disadvantage of this approach is that the parameters may contain measurement errors from the indicator function that were not taken into account during the choice model. 

To solve this issue, another approach uses simultaneous estimation of structural and measurement model, which includes the latent variable in the choice model framework. This is so called the Integrated Choice and Latent Variable (ICLV) model (Fig. \ref{icmc_fig00}b). The ICLV model explicitly uses information from measurement indicators and explanatory variables to derive latent constructs. This combined structural model framework has led to many interesting results, e.g. environmental attitudes in rail travel \citep{hess2013accommodating},  image, stress and safety attitudes towards cycling \citep{maldonado2014exploring}, and social attitudes towards electric cars \citep{kim2014expanding}. However, the simultaneous approach still relies on a separate measurement model (latent variable model) that describes the relationship to indicators. Despite the direct benefits of the ICLV model combining factor analysis with traditional discrete choice models, the only advantage to using such an approach is when attitudinal measurement indicators are expected to be available to the modeller and the observed explanatory variables are weak predictors of the choice model \citep{vij2016and}. Even when measurement indicators are available, they may not provide any further information that directly influence the choice than through explanatory variables \citep{chorus2014possibility}. Consequently, mis-specification and other measurement errors may occur, when the criteria is not associated with the choice model. Without measurement indicators to guide selection of latent variables, we can alternatively use ML for latent variables through data mining. This can be implemented through generative modelling methods used in ML. Generative modelling in ML is a class of models which uses unlabelled data to generate latent features. Generative models learn the underlying choice distribution $p(y)$ and the latent inference $p(h|y)$, where $h$ is the latent variable. Followed by implementing a Bayesian network that represents a probabilistic conditional relationship between random variables and dependencies to derive the posterior distribution of $y$ given $h$ using $p(y|h) = \frac{p(h|y)p(y)}{p(h)}$. Efficient algorithms which perform ML and inference such as RBMs can be used in this method. The denominator is given by $p(h) = \sum_y p(h|y=1)$ indicating choice $y$ is chosen. The rapid advancement of machine learning research have led to the development of efficient semi-supervised training algorithms such as the conditional restricted Boltzmann machine (C-RBM) \citep{salakhutdinov2007restricted,larochelle2008classification}, a hybrid discriminative-generative model, capable of simultaneously estimating a latent variable model using a priori choice distribution with an latent inference model (see Fig. \ref{icmc_fig01}).

To date, econometric and machine learning models are often studied for its contrasting purposes in decision forecasting by behavioural researchers \citep{breiman2001statistical}. Econometric models are based on the classical decision theory that individual’s decisions can be modelled rationally based on utility maximization. These models assume that the population will adhere to the strict formulation of the choice model, but may not always represent the true decisions. Generative modelling based approach uses clustering and factor analysis developed through algorithmic modelling of the data. Associations between decision factors can be classified in this method, obtaining latent information without explicit definition of latent constructs \citep{poucin2016pedestrian}. Thus, machine learning algorithms such as ANN that decouple latent information from `true' distribution generally outperform traditional regression based models in multidimensional problems \citep{ahmed2010empirical}. Recent works on latent behaviour modelling on choice analysis agree on the potential of improving behaviour models with machine learning. Examples include combining machine learning to improve complex psychological models \citep{rosenfeld2012combining}, representing the phenomena of similarity, attraction and compromise in choice models \citep{osogami2014restricted} and inference of priorities and attitudinal characteristics \citep{aggarwal2016learning}.

Despite the many benefits, interpretation of results are still extremely difficult due to the complexity and number of parameters in ML analysis. As a result, ML models are not often used for general purpose behaviour understanding, but created exclusively for a specific purpose for prediction accuracy. Still, machine learning research is a rapidly growing field at the intersection of statistical analysis and information science to find patterns in complex data \citep{donoho201550}. Furthermore, with the emphasis on applications and theoretical studies in today's massive data driven industry, improving analytical techniques with ML is very relevant, although structural modelling, statistical and probability theory will remain the cornerstone of discrete choice analysis.

\subsection{The basis of latent class and latent variable models}
\begin{figure}[!b]
	\centering
	\includegraphics[width=0.95\linewidth]{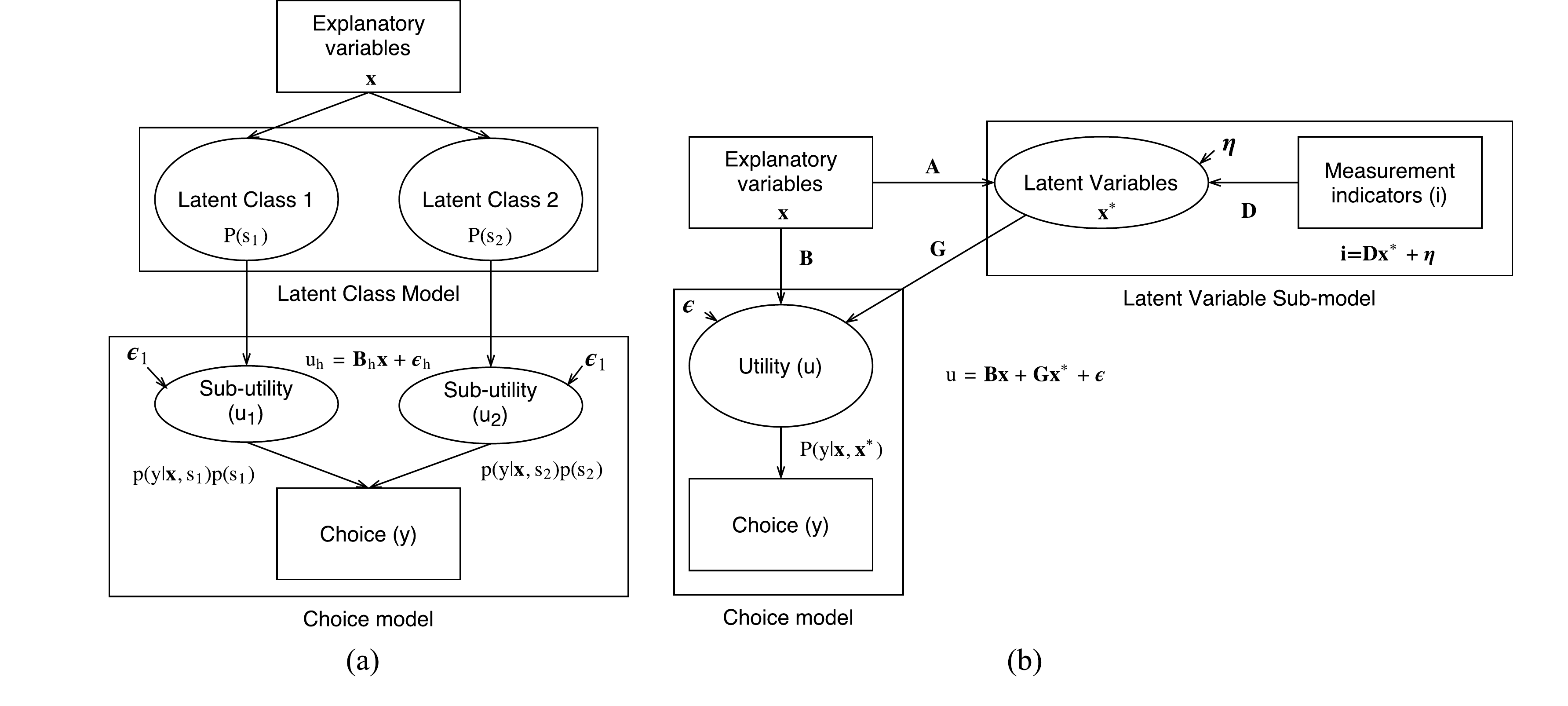}
	\caption{Classical structural framework for (a) latent class model and (b) integrated choice and latent variables model}
	\label{icmc_fig00}
\end{figure}

The latent class model shown in Figure \ref{icmc_fig00} is a simple top-down model that imparts generalization properties to the choice model that predefines a discrete number of classes, allowing the parameters to vary with an fixed distribution. Formally, the LCM choice probability can be expressed as:

\begin{equation}
P(y) = \sum\limits_n P(s_n)P(y|x, s_n)
\end{equation}

where $S=[s_1, s_2,...,s_n]$ are the set of classes and $P(s_n)$ is the probability that an individual belongs to class $s$. $P(y|x, s_n)$ is the conditional probability of choice $y$ selected given the class $s_n$ and input variable $x$.

\begin{figure}
	\centering
	\includegraphics[width=0.67\linewidth]{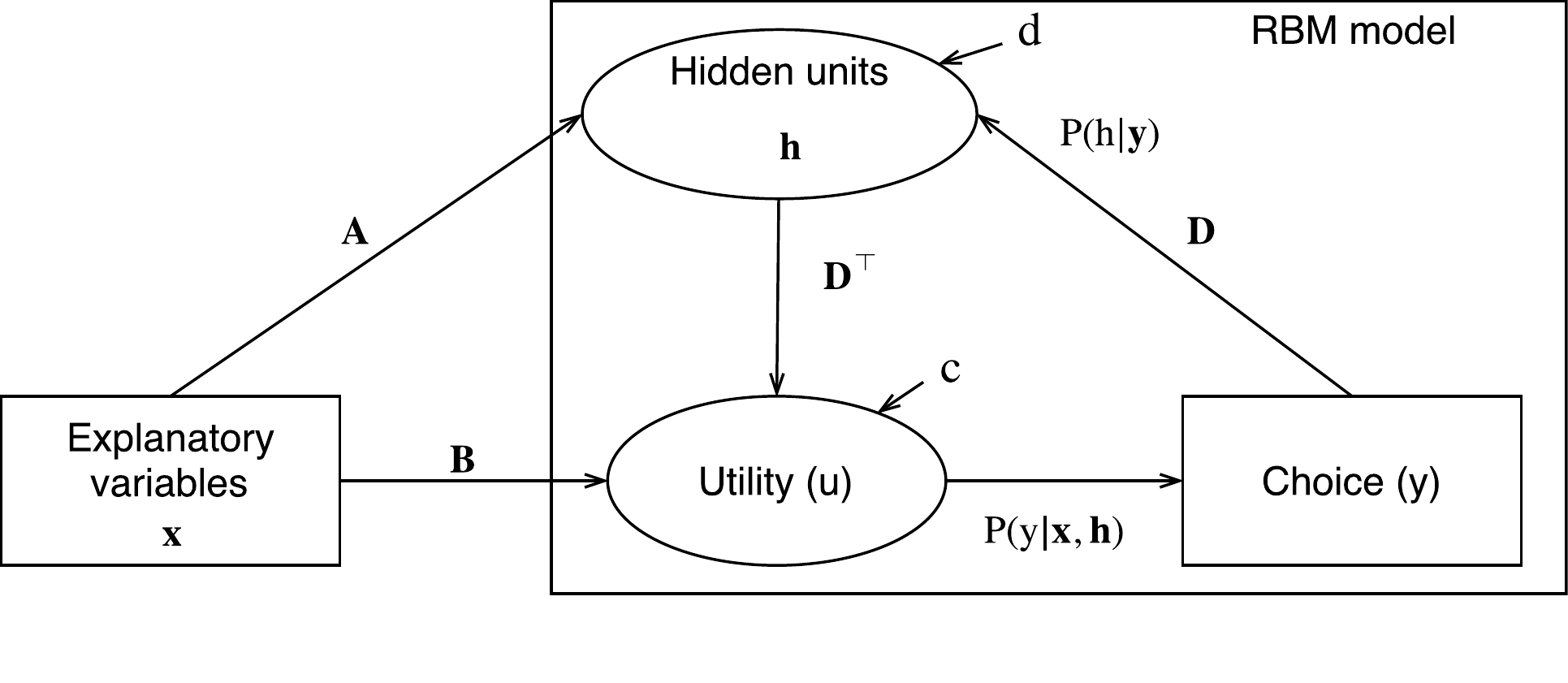}
	\caption{Framework for a C-RBM choice model conditional on explanatory variables and choice distribution.}
	\label{icmc_fig01}
\end{figure}

The ICLV model extends the choice model by describing how perceptions and attitudes affect real choices as well as using separate indicators to estimate latent variables \citep{ben2002hybrid}. Latent variables can be classified as either attitudinal (individual characteristics) or perceived (personal beliefs towards responses) \citep{ben1999discrete}. The latent variable model (measurement model) forms a sub-part of the structural framework which captures the relationship between the latent variables and indicators and the observed explanatory variables which influence the latent variables. This specification can be used to identify more useful parameters and predict accurate decision outcomes when there is a lack of strong significant correlation between explanatory variables and choice outcomes. The functions of the structural and measurement model can be explained in four equations \citep{vij2016and}:

\begin{equation}
\label{eq_lv00}
\textbf{x}^*=\textbf{Ax} + \bm{\nu}
\end{equation} 
\begin{equation}
\label{eq_lv01}
\textbf{I}^*=\textbf{Dx}^* + \bm{\eta}
\end{equation}
\begin{equation}
\textbf{u}=\textbf{Bx} + \textbf{Gx}^* + \bm{\epsilon}
\end{equation}
\begin{equation}
y_i=\begin{cases}
1 \text{  if  } u_i>u_i' \text{ for } i \in \lbrace 1,...,I \rbrace\\ 
0 \text{  otherwise  } \\
\end{cases}
\end{equation}

where $u_i$ is the utility of selecting alternative $i$. $\textbf{A}$ represents the relationship between input explanatory variables $\textbf{x}$ and latent variables $\textbf{x}^*$, $\textbf{D}$ represents the relationship between $\textbf{x}$ and the indicator output $\mathbf{I}^*$. $\textbf{B}$ and $\textbf{G}$ represents the model parameters with respect to the observed and latent variables. $\bm{\nu}$, $\bm{\eta}$ and $\bm{\epsilon}$ are the stochastic error terms of the model, assumed to be mutually independent and Gumbel distributed. In a generative model, parameters are shared between $\textbf{G}$ and $\textbf{D}$ that simply defines the joint distribution of $p(y,h)$, i.e. $\textbf{G} = \textbf{D}^\top$ (Fig. \ref{icmc_fig01}). The re-use of a shared parameter vector differentiates the RBM model from the structural equation formulation of the ICLV model.

\subsection{Modelling through generative machine learning methods}
In generative machine learning models, hidden units $h$ are the learned features (see Fig. \ref{icmc_fig01}) which performs non-redundant generalization of the data to reduce high dimensional input data \citep{hinton2006reducing}. Intuitively, in terms of econometric analysis, hidden units are latent variables that depend on some observed data, for instance, socio-economic attributes such as weather or price information or direct choices such as location and choice of purchase.  We can construct a generative model as a function of these dependent and independent variables. In the case of factor analysis approach, a common process is to perform feature extraction based on statistical hypothesis testing to determine if the values of the two classes are distinct, for example, using Support Vector Machines (SVMs) or Principal Component Analysis (PCA) to learn low-dimensional classes by capturing only significant statistical variances in the data \citep{poucin2016pedestrian,wong2016bike}. The learned classes (or clusters) can then be introduced directly into the model via parameterization. In generative modelling approach, we use the priors directly to learn the distribution of the hidden units. In this process we extract latent information directly from the observed choice data instead of using measurement functions which may be prone to errors.

\subsection{Balancing model inference and accuracy}
One common problem that researchers face when constructing latent behaviour models is specifying of the optimal size of latent factors \citep{vermunt2002latent}. Since the hypothesis on the number of latent size cannot be tested directly, typical statistical evaluation methods such as AIC and BIC are used to guide class selection \citep{vermunt2002latent}, in the case of ICLV models, through predefinition of measurement functions \citep{Rungie2012145}. However, since the number of latent factors determines the ability of the model to represent the various heterogeneity in the data, it is likely that as we increase $h$, the choice model become more efficient in capturing complex behaviour effects from individual and latent attributes. On the other hand, if we increase the number of latent segments, the number of parameters will also increase at an exponential rate \citep{vermunt2002latent}. Therefore, we may gain model accuracy but we would lose model interpretability. 

The trade-off between inference and accuracy is a challenge when dealing with complex data \citep{breiman2001statistical}. If the goal of latent behaviour modelling is to leverage on data to understand underlying statistical problems, we have to incorporate implicit modelling methods in addition to describing explicit structural utility formulations.

\section{Methodology}
\label{S:3}
In this section, we provide a brief overview on restricted Boltzmann machines and how it can be used to generate prior over the choice distributions. We refer readers to \cite{Goodfellow-et-al-2016} for background and details on generative models and deep learning.

\subsection{Restricted Boltzmann machines}
A restricted Boltzmann machine (RBM) is an energy-based \textit{undirected graphical model} that extends from a Markov Random Field distribution by including hidden variables \citep{salakhutdinov2007restricted}. It is a single layer artificial neural network with no internal layer connections. The model has stochastic visible variables $\mathbf{y} \in \lbrace 0,1 \rbrace^I$ and stochastic hidden variables $\mathbf{h} \in \lbrace 0,1 \rbrace^J$. The joint configuration $ (\mathbf{y, h}) $ of visible and hidden variables is given by the Hopfield energy \citep{hinton1984boltzmann}:

\begin{equation}
\text{Energy}(\mathbf{y, h}) = -\sum\limits_{i \in vis} y_i c_i -\sum\limits_{j \in hid} h_j d_j -\sum\limits_{i,j} h_j D_{ij} y_i,
\end{equation}

where $d_j$ and $c_i$ represent the vector biases (constants) for the hidden and visible vectors respectively. $D_{ij}$ is the matrix of parameters representing an undirected connection between the hidden and visible variables. We can express the Boltzmann distribution as an energy model with energy function $F(\mathbf{y})$: 

\begin{equation}
p(\mathbf{y, h}) = \frac{1}{Z} \exp(-F(\mathbf{y})),
\end{equation}

where the partition function $Z=\sum\limits_{i, j} \exp(-\text{Energy}(\mathbf{y, h}))$ is the normalization function over all possible vector combinations. $F(\mathbf{y})$ is defined as the free energy $F(\mathbf{y}) = -\ln \sum_h \exp(-\text{Energy}(\mathbf{y, h}))$ and further simplified to 

\begin{equation}
F(\mathbf{y}) = -y_i c_i - \sum\limits_{j \in hid} \ln (1+\exp(D_{.,j} y + d_j)).
\end{equation}

The probability of assigning a visible vector $ \mathbf{y} $ is given by the sum of all possible hidden vector states:

\begin{equation}
p(\mathbf{y}) = \frac{1}{Z} \sum\limits_{\mathbf{h}}\exp(-F(\mathbf{y})).
\end{equation}

The RBM model is used to learn aspects of an unknown probability distribution based on samples from that distribution. Given some observation, the RBM makes updates to the model weights such that the model best represent the distribution of the observation. To generate data with this method, it is necessary to compute the log likelihood gradient for all visible and hidden units. Hinton introduced a fast greedy algorithm to learn model parameters efficiently using Contrastive Divergence (CD) method that starts a sampling chain (Gibbs sampling) from real data points instead of random initialization \citep{hinton2010practical}.

\subsection{Model estimation and inference}
The probability that the RBM network learns a training sample can be raised by adjusting the weights to lower the energy of that training sample and raise the energy of other non-training samples. In order to minimize the negative log likelihood of the probability distribution $p(\mathbf{y})$, we take its gradient derivative of the log probability of a training vector with respect to the model parameters as follows:

\begin{equation}
\frac{\partial \log p(\mathbf{y})}{\partial \theta} = \langle y_i h_j\rangle_{train} - \langle y_i h_j \rangle_{model} = \phi^+ -\phi^-,
\end{equation}

where the components in the angle brackets corresponds to the expectations under the specified distribution. The first and second terms are the positive $\phi^+$ and negative $\phi^-$ phases respectively. This function updates the model parameters using a simple learning rule with a learning rate $ \Phi $:

\begin{equation}
\Delta \theta = \Phi (\langle y_i h_j\rangle_{train} - \langle y_i h_j \rangle_{model}).
\end{equation}

The updates for parameters $\theta=\lbrace D_{ij}, d_j, c_i \rbrace$ can be performed using simple stochastic gradient descent at each iteration of $t$:

\begin{equation}
\theta_{t} = \theta_{t-1} - \Delta \theta.
\end{equation}

To obtain a sample of a hidden unit from $\langle y_i h_j\rangle_{train}$, we take a random training sample $\textbf{y}$ and sample the state in the hidden layer is given by the following function:

\begin{equation}
\label{eq7}
p(h_j=1|\mathbf{y}) = \frac{e^{d_j+\sum_i D_{ij} y_i}}{1+e^{d_j + \sum_i W_{ij} y_i}} = \sigma(d_j + \sum\limits_i D_{ij} y_i),
\end{equation}

where $\sigma(x)={e^x}/{(1+e^x)}$. Similarly, we can obtain a visible state, given a vector of sampled hidden units, via a logistic function:

\begin{equation}
\label{eq8}
p(y_i|\mathbf{h}) = \frac{e^{c_i + \sum_j D_{ij} h_j}}{\sum_i e^{c_{i'} + \sum_j D_{i'j} h_j}}.
\end{equation}

Since weights are shared between $D$ and $G$ and they define the distributions of $p(y), p(h), p(y,h), p(y|h)$ and $p(h|y)$, we can express the posterior distribution as $p(y) = \sum_h p(h)p(y|h)$ \citep{ng2002discriminative}. Due to its bidirectional structure, this framework possesses good generalization capabilities. The visible layer represents the data (in the case of choice modelling, data represent selected choices), and the hidden layer represents the capacity of the model as class distributions.

The model can be inferred from $\langle y_i h_j \rangle_{model}$ can be done by setting the states of the visible variables to a training sample and then the states of the hidden variables are computed using Eq. \ref{eq7}. Once a ``state'' is chosen for the hidden variables, a ``reconstruction'' phase produces a new vector $\tilde{\textbf{y}}$ with a probability given by Eq. \ref{eq8}, and the gradient update rule is given by:

\begin{equation}
\Delta \theta = \Phi (\langle y_i h_j\rangle_{train} - \langle y_i h_j \rangle_{reconstruction}).
\end{equation}

We approximate the gradient function by using a CD Gibbs sampler minimizing the divergence between the expected and estimated probability distribution, known as the Kullback-Leibler (KL) divergence \citep{hinton2002training}. A divergence ratio of 0 indicates that the estimated distribution is totally similar. The training algorithm that runs for a total number of $N$ chain steps is initialized from a fixed point from the data distribution and then averaged across all examples \citep{carreira2005contrastive}.

\subsection{Modelling approach}
\label{S:4}
In this paper, the proposed method uses a conditional RBM (C-RBM) training algorithm to include input-output connections that allows for discriminative learning \citep{mnih2012conditional}. C-RBM expands the model to include ``context input variables'', i.e. $p(y|x,h)$. $k$ input explanatory variables are introduced as context variables so that they can be used to influence the latent variables, even though Eq.\ref{eq8} does not reconstruct these explanatory variables. This influence is represented by a weight matrix $B_{ik}$. The intuition is that for each latent variable, it acts as a function of the observed choice $\textbf{y}$, conditional on $\textbf{x}$ (see Fig. \ref{icmc_fig01}). In the choice prediction stage, a vector of new input samples $\textbf{x}$ generate latent variables $\textbf{h}$. Conditional on the explanatory and latent variables, a probability function describing the choice behaviour is given as:

\begin{equation}
\label{eq9}
p(y_i|\mathbf{h,x}) = \frac{e^{\sum_k B_{ik} x_k + \sum_j D_{ij} h_j + c_i}}{\sum_{i'} e^{\sum_k B_{i'k} x_k + \sum_j D_{i'j} h_j + c_{i'}}}.
\end{equation}

\begin{figure}
	\centering
	\includegraphics[width=\textwidth]{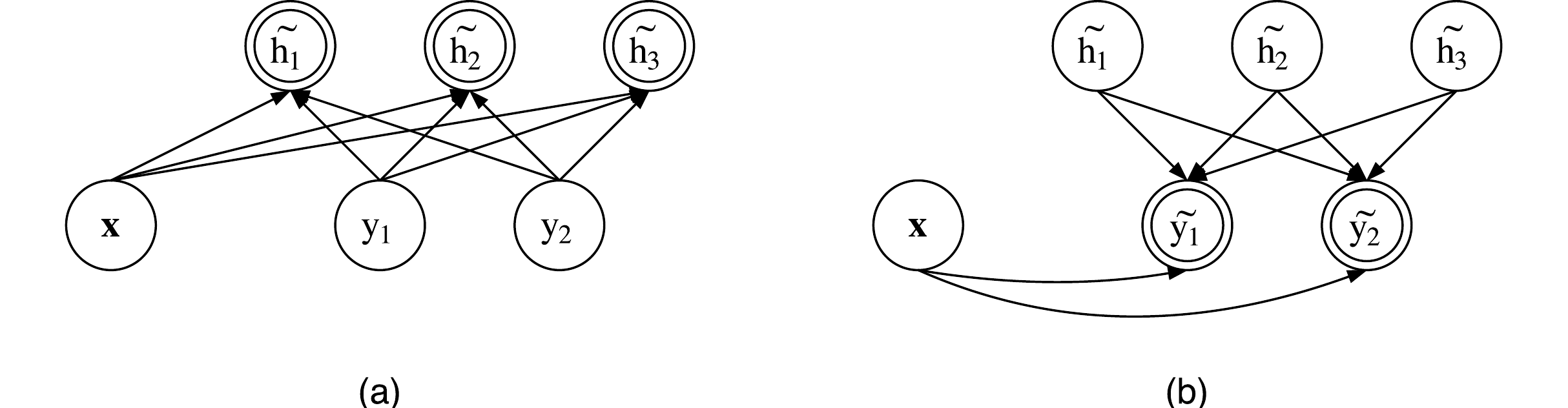}
	\caption{C-RBM (a) positive $\phi+$ and (b) negative $\phi-$ phases during semi-supervised discriminative training. Weights (connections) are learned to reduce reconstruction $\tilde{y}$ error.}
	\label{icmc_fig02}
\end{figure}

\begin{figure}
	\centering
	\includegraphics[width=\textwidth]{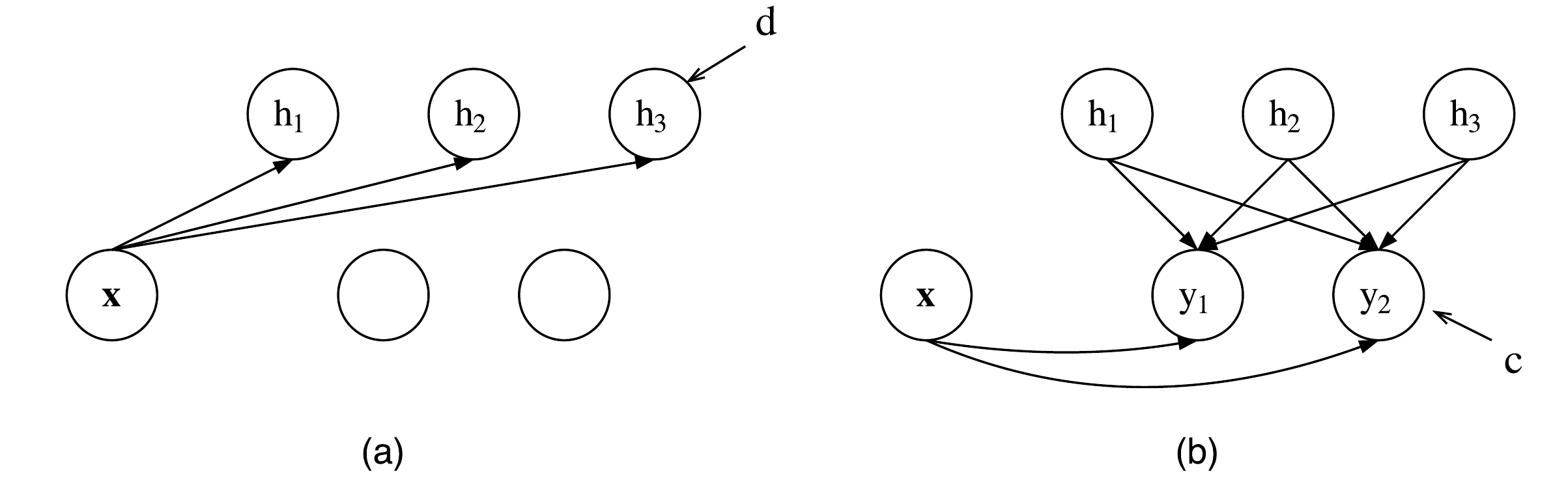}
	\caption{During the choice prediction phase, (a) latent variables are sampled using explanatory variables, and (b) the choice model is estimated with variables $x$ and $h$.}
	\label{icmc_fig04}
\end{figure}

Likewise, sampling of the hidden state is extended to incorporate $\mathbf{x}$:

\begin{equation}
\label{eq7.1}
p(h_j=1|\mathbf{y}) = \sigma(d_j + \sum\limits_i D_{ij} y_i + \sum\limits_k A_{jk} x_k),
\end{equation}

where the update parameters are $\theta=\lbrace D_{ij}, B_{ik}, A_{jk},d_j, c_i \rbrace$. During the reconstruction phase, the condition probability (Eq. \ref{eq9}) is equivalent to a MNL model with latent variables (where $h$ and $x$ represents the latent and observed variables respectively). Good latent variables $h$ best capture information along the orthogonal direction where choices $y$ and observed inputs $x$ vary the most. The training and choice estimation phase is illustrated in Fig. \ref{icmc_fig02} and \ref{icmc_fig04}. In the positive phase, parameter vectors are adjusted decided by the learning rate $\sigma$ to learn the transformed latent representation of the training set. In the negative phase, the latent variables are ``clamped'' or realized and the parameter vectors are adjusted again by reconstructing the observed variables. Referring to Fig. \ref{icmc_fig01}, the multinomial (MNL) model estimates the conditional parameter vector $\textbf{B}$ and bias vector $\bm{c}$, while the C-RBM model includes vectors $\textbf{D}$, $\textbf{A}$ and $\bm{d}$.

\section{Data}
\label{S:4}
In this section, we develop a financial product choice scenario with explanatory variables using the C-RBM model. The latent variables representing the latent attitudinal variables is simultaneously estimated in conjunction with the interaction with choice model. First, we construct a structured choice subset from a financial product transaction dataset from the Kaggle database\footnote{Dataset: https://www.kaggle.com/c/santander-product-recommendation/data}. The data shows a monthly basis record of each financial product purchase by customers of Santander. The time span of the data is from January 2015 to June 2016. Next, we reduced the complexity of the dataset by removing transaction data which contain multiple product choices. To ensure consistency, inputs were scaled and normalized. Overall, the constructed dataset has a total of 13 alternatives (product choice) and 20 explanatory variables. Table \ref{table_choix} lists the alternatives and distribution across the dataset. Given the above conditions, a total of 253,803 valid responses were recorded representing the total population sample with 13 available choices. A descriptive list of mean and standard deviation values of the explanatory variables are shown in Table \ref{table00}. The experimental question is straightforward: ``Given a set of examples with explanatory variables, what product is the individual most likely to purchase in the given month?'' In a typical situation, the decision maker chooses an alternative that yields the maximum utility, making an inference about the behaviour of the decision maker using the predictive model.

\begin{table}[!ht]
	\centering
	\caption{Choices ($\textbf{y}$)}
	\label{table_choix}
	\begin{tabular}{l l l}
		\toprule
		Choice index & Name & Total sample distrib.\\
		\midrule
		1 & Guarantees & 0.002\%\\
		2 & Short-term deposits & 0.83\%\\
		3 & Medium-term deposits & 0.07\%\\
		4 & Long-term deposits & 3.79\%\\
		5 & Funds & 0.98\%\\
		6 & Mortgage & 0.02\%\\
		7 & Pensions &0.15\%\\
		8 & Loans &0.035\%\\
		9 & Taxes &2.68\%\\
		10 & Cards &21.93\%\\
		11 & Securities &1.42\%\\
		12 & Payroll &22.04\%\\
		13 & Direct debit &46.05\%\\
		\bottomrule
	\end{tabular}
\end{table}

\begin{table}[ht]
	\caption{Explanatory variable descriptive statistics ($\textbf{x}$)}
	\label{table00}
	\centering
	\begin{tabular}{l p{6.9cm} l l}
		\toprule
		Explanatory variable & Description & mean & std. dev.\\
		\midrule
		age & Customer age & 42.9 & 13.0\\
		loyalty & Customer seniority (in years) & 8.03 & 6.0\\
		income & Customer income (\euro{}) & 141,838 & 262,748\\
		sex & Customer sex (1=male) & 0.387 & 0.487\\
		employee & Employee index, 1 if employee & 0.0006 & 0.024\\
		active & Active customer index & 0.95 & 0.199\\
		new\_cust & 1 if customer loyalty $<$ 6 mo. & 0.045 & 0.207\\
		resident & Resident index (Spain) & 0.999 & 0.007\\
		foreigner & Foreign citizen index & 0.045 & 0.21\\
		european & EU citizen index & 0.995 & 0.006\\
		vip & VIP customer index & 0.116 & 0.32\\
		savings & \textit{Savings} Account type & 0.0002 & 0.012\\
		current & \textit{Current} Account type & 0.572 & 0.495\\
		derivada & \textit{Derivada} Account type & 0.0009 & 0.03\\
		payroll\_acc & \textit{Payroll} Account type & 0.416 & 0.493\\
		junior & \textit{Junior} Account type & 0.0001 & 0.0098\\
		masparti & \textit{Mas Particular} Account type & 0.017 & 0.128\\
		particular & \textit{Particular} Account type & 0.168 & 0.373\\
		partiplus & \textit{Particular Plus} Account type & 0.113 & 0.316\\
		e\_acc & \textit{e-Account} type & 0.255 & 0.436\\
		\bottomrule
	\end{tabular}
\end{table}

\subsection{Method for assessing C-RBM model performance}
We can estimate the weights for the latent inference model $ B_{ik} $ and $ D_{ij} $ by optimizing the lower bound of the KL-divergence using gradient backpropagation. Intuitively $ D_{ij} $ represents the parameters for the explanatory variables and $ B_{ik} $ represents the parameters for the latent variables. We selected models with 2, 4, 16 and 32 latent variables to observe the effects of increasing model complexity. One disadvantage of this step is that it results in a large number of estimated parameters: $(N_{params} \in \mathbb{R}^{(I\times J) + (K \times I) + (K \times  J) + K + I})$. With $J = 4$, we ended up with 409 parameters. To counteract overfitting due to this problem, we trained on 70\% of our data and validate the model on the other 30\% with a 2-fold cross-validation to verify generalization. When the validation error stops decreasing, the optimal estimation is reached \citep{Goodfellow-et-al-2016}. A baseline comparison is set up using a standard multinomial logistic regression model with all explanatory variable and compared to the discriminative C-RBM modelling approach, followed by comparing the log-likelihood, $\rho^2$ model fit and predictive accuracy across all data models. The criteria for measuring performance of a categorical based model include: $\rho^2$ model fit and prediction error. The $\rho^2$ fit denotes the predictive ability between the trained model and a model without covariates. In the prediction error evaluation, the elements in the diagonal cells of a confusion matrix over the total number of examples denotes the accuracy of the model in predicting the correct choice and the error is

\begin{equation}
\text{Error}_{valid} = 1 - \sum_i P(y_{pred} = 1 | x, h, y_i = 1).
\end{equation}

$y_i$ is the actual choice and $\text{Error}_{valid}$ is the sum of all the error probabilities for correct assessment for each choice. We fit the model on the training set and evaluate on the validation set.

\section{Results}
\label{S:5}
We compare the different models based on their generalization performance on the test set. A total of 76,141 observations were used in the test. For the purpose of this study, we tested on both normalized and non-normalized data and found that both data produce similar result. Model estimation and validation were performed with Theano ML Python libraries\footnote{Theano Python library: http://github.com/Theano/Theano}. Optimization parameters used were stochastic gradient descent (SGD) on mini-batches of 64 samples for 400 epochs with input normalization. We used an adaptive momentum based learning rate of with initial rate of $1e^-3$ \citep{hinton2006fast}. Training time was approximately 30 minutes for each model including validation running on a Intel Core i5 workstation. At the given time, computational demand may not be significant to justify the small number of hidden units, however, speed could become a more important consideration when model estimation and validation increase in data size or using very large parameter vectors with higher dimensionality. The statistical results of the model comparison across the same validation set is shown in Table  \ref{table_modelstats}. We found that additional latent information about the relationship between explanatory variables and observed decisions was useful and increases model accuracy. Bayesian Information Criterion (BIC) values indicate that 8 hidden units may be the optimal number of latent variables and higher BIC values above 8 hidden units might suggest overfitting. However, when generating semantic class meanings, a smaller number of latent variables may be simpler, therefore, in our example, we use only 2 latent variables for analysis. 

\begin{figure}[ht]
	\centering
	\includegraphics[width=0.6\textwidth]{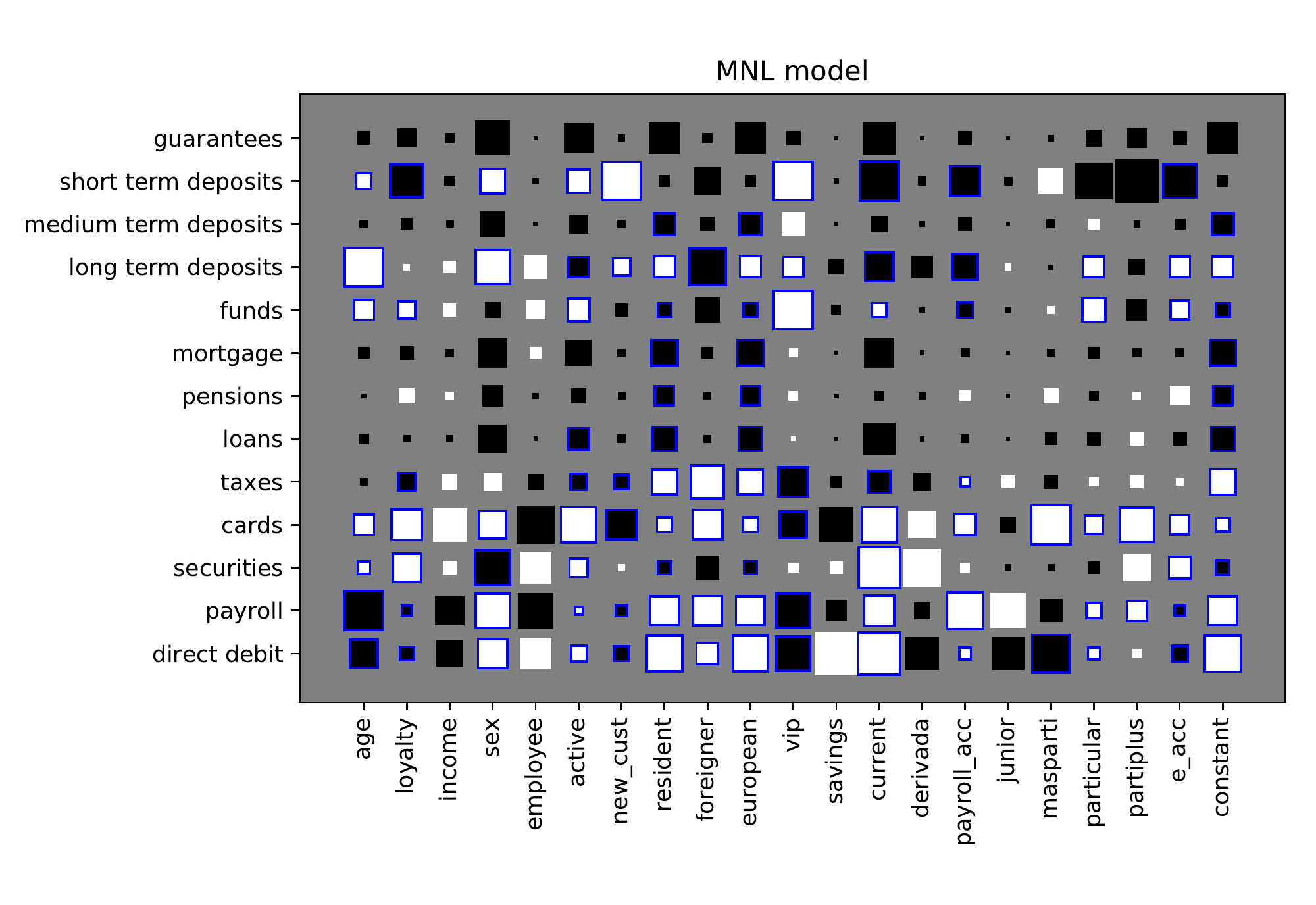}
	\caption{MNL model parameters. White: +ve values, Black: -ve values, Blue: >95\% significant}
	\label{matrix_00}
\end{figure}

\begin{figure}[ht]
	\makebox[\linewidth][c]{
		\begin{subfigure}{0.6\textwidth}
			\includegraphics[width=1\textwidth]{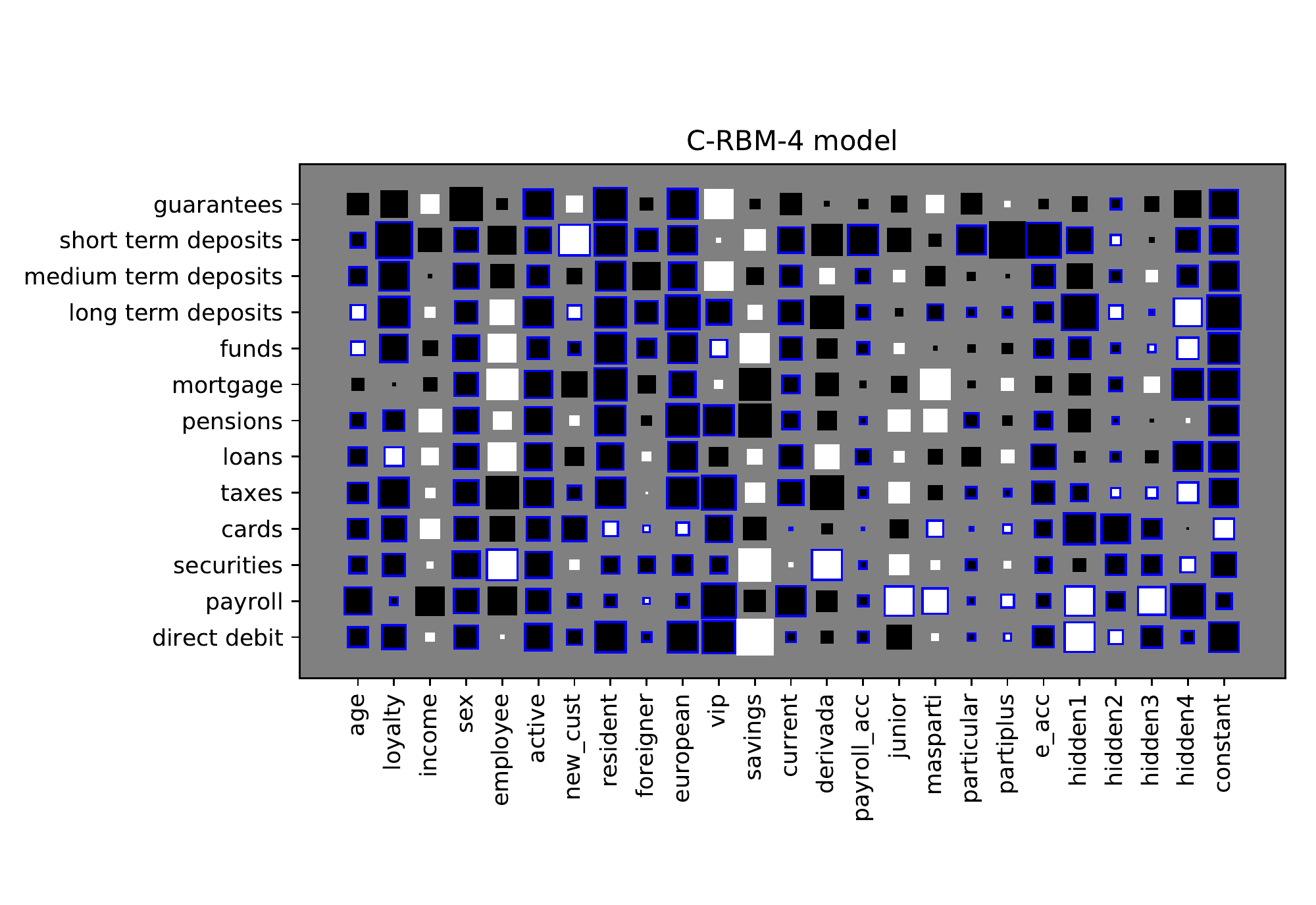}
			\caption{}
			\label{matrix_03}
		\end{subfigure}
		\begin{subfigure}{0.6\textwidth}
			\includegraphics[width=1\textwidth]{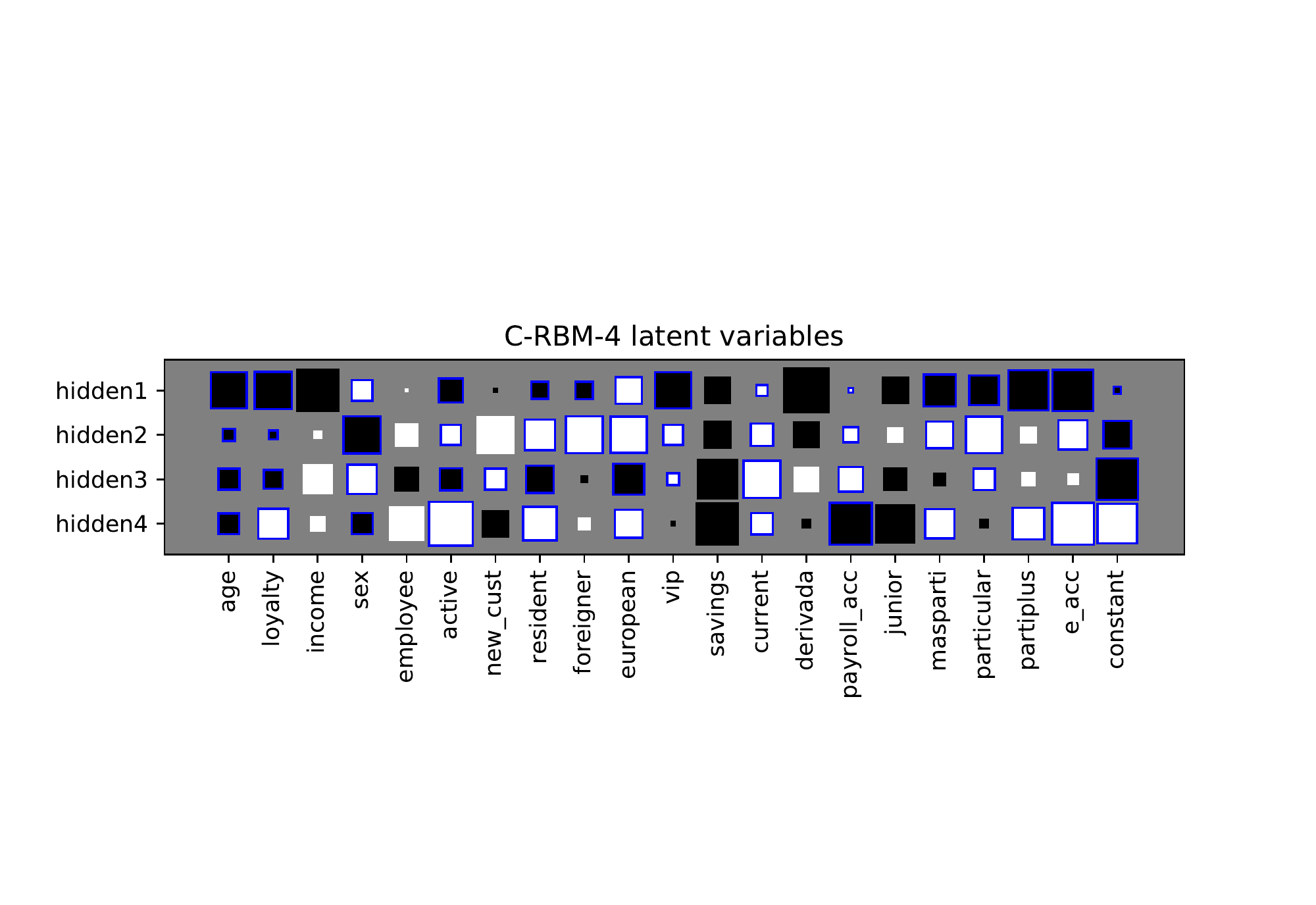}
			\caption{}
			\label{matrix_04}
		\end{subfigure}
	}
	\caption{(a) C-RBM model with 4 latent variables. (b) Latent variable relationship parameters. White: +ve values, Black: -ve values, Blue: >95\% significant}
	\label{fig_crbm4}
\end{figure}

\begin{figure}
	\makebox[\linewidth][c]{
		\begin{subfigure}{0.6\textwidth}
			\includegraphics[width=1\textwidth]{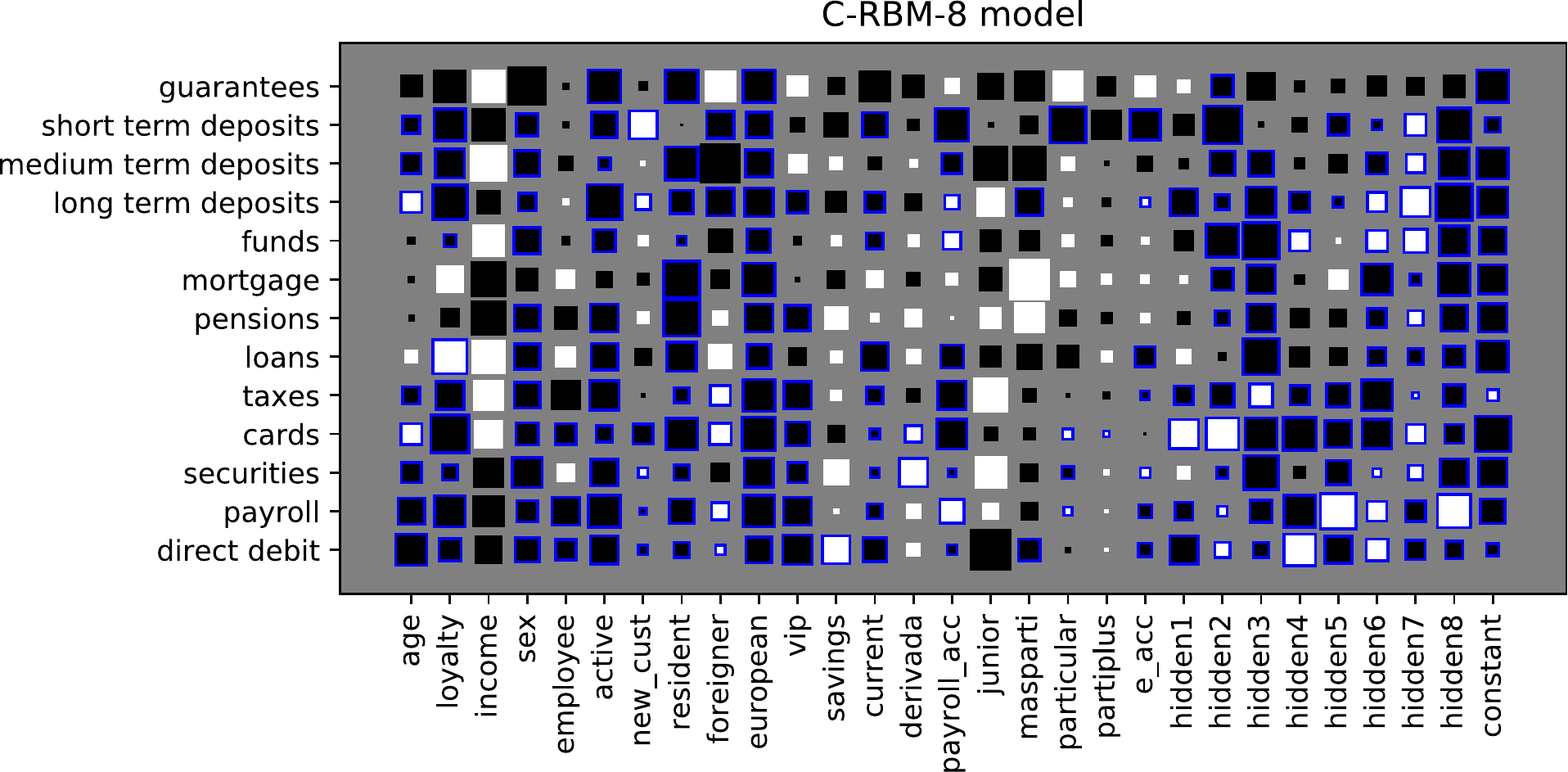}
			\caption{}
			\label{matrix_05}
		\end{subfigure}
		\begin{subfigure}{0.6\textwidth}
			\includegraphics[width=1\textwidth]{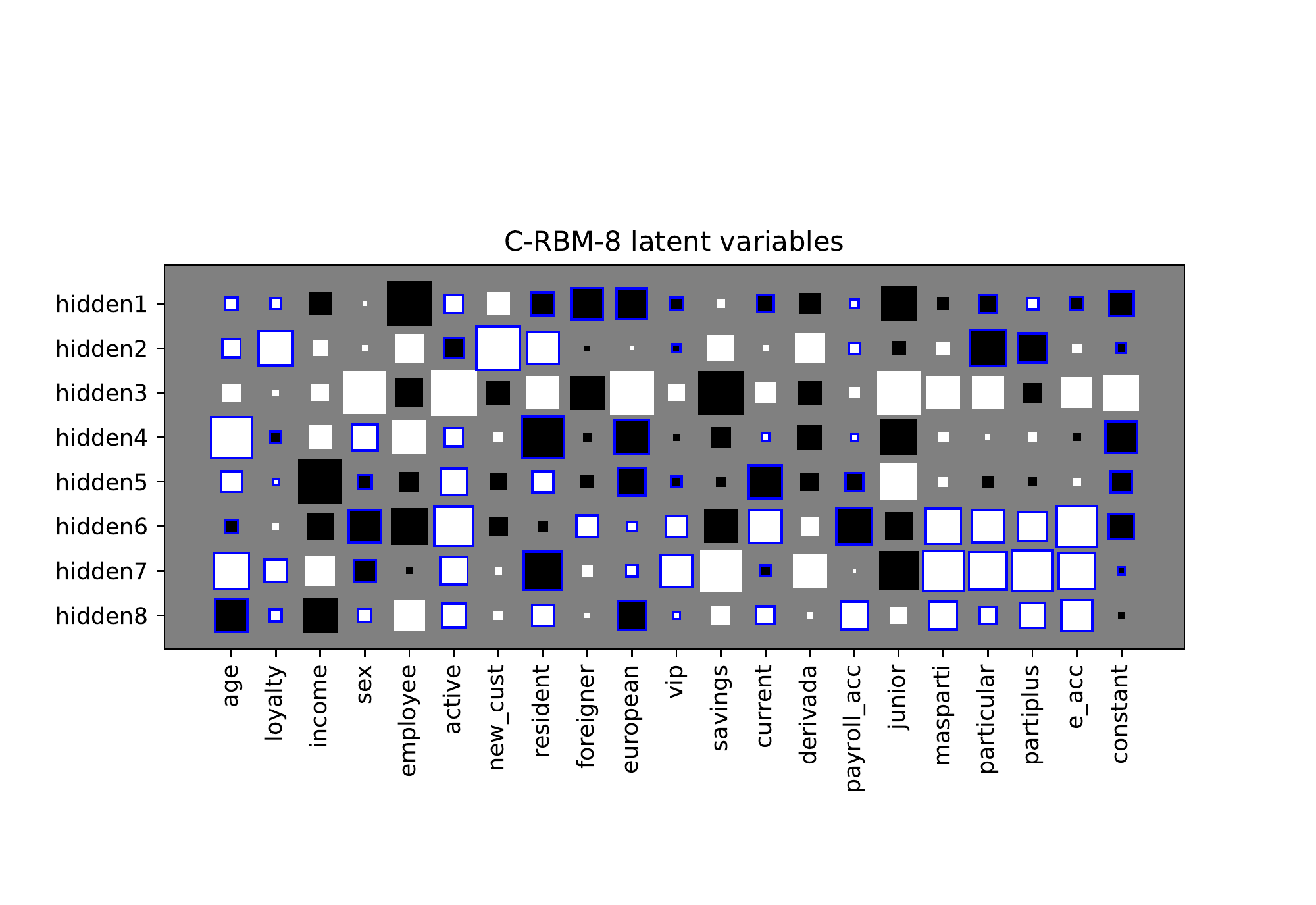}
			\caption{}
			\label{matrix_06}
		\end{subfigure}
	}
	\caption{(a) C-RBM model with 8 latent variables. (b) Latent variable relationship parameters. White: +ve values, Black: -ve values, Blue: >95\% significant}
	\label{fig_crbm8}
\end{figure}

\begin{figure}[ht]
	\makebox[\linewidth][c]{
		\begin{subfigure}{0.6\textwidth}
			\includegraphics[width=1\textwidth]{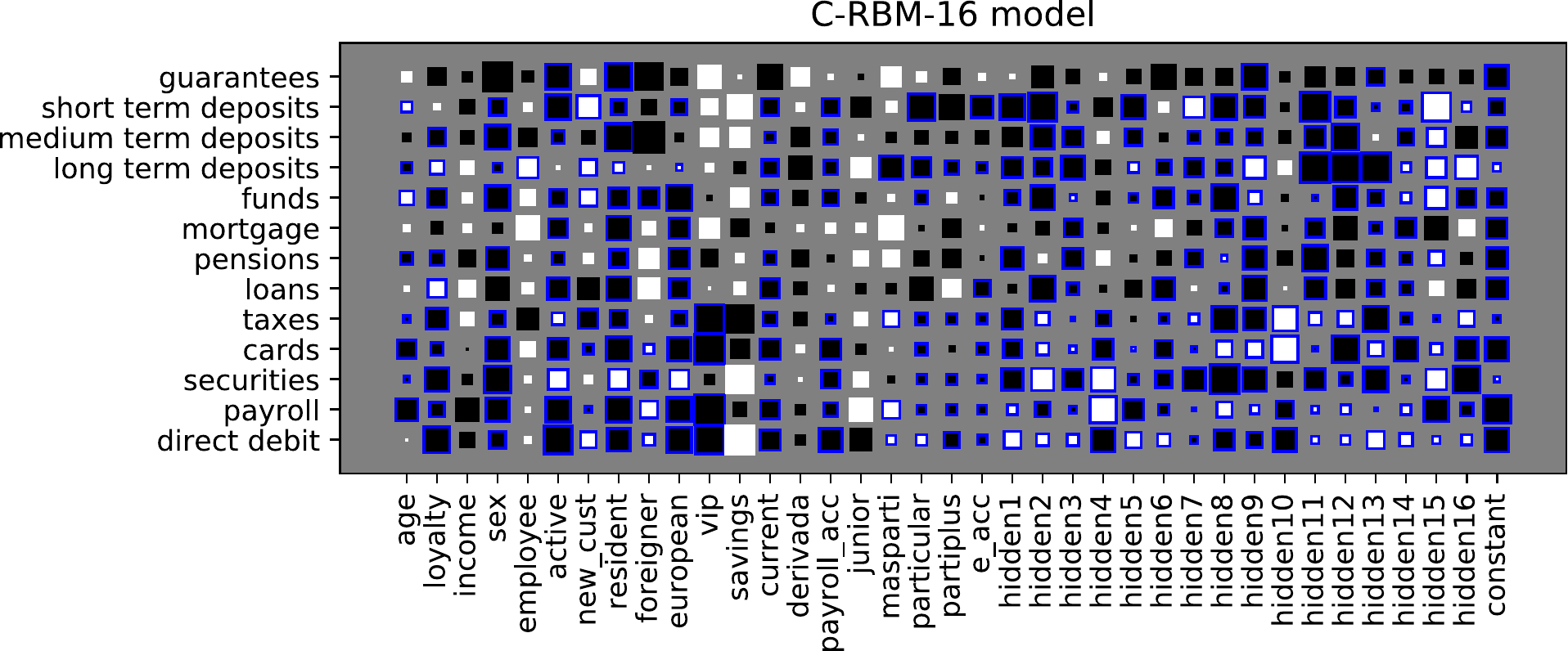}
			\caption{}
			\label{matrix_07}
		\end{subfigure}
		\begin{subfigure}{0.6\textwidth}
			\includegraphics[width=1\textwidth]{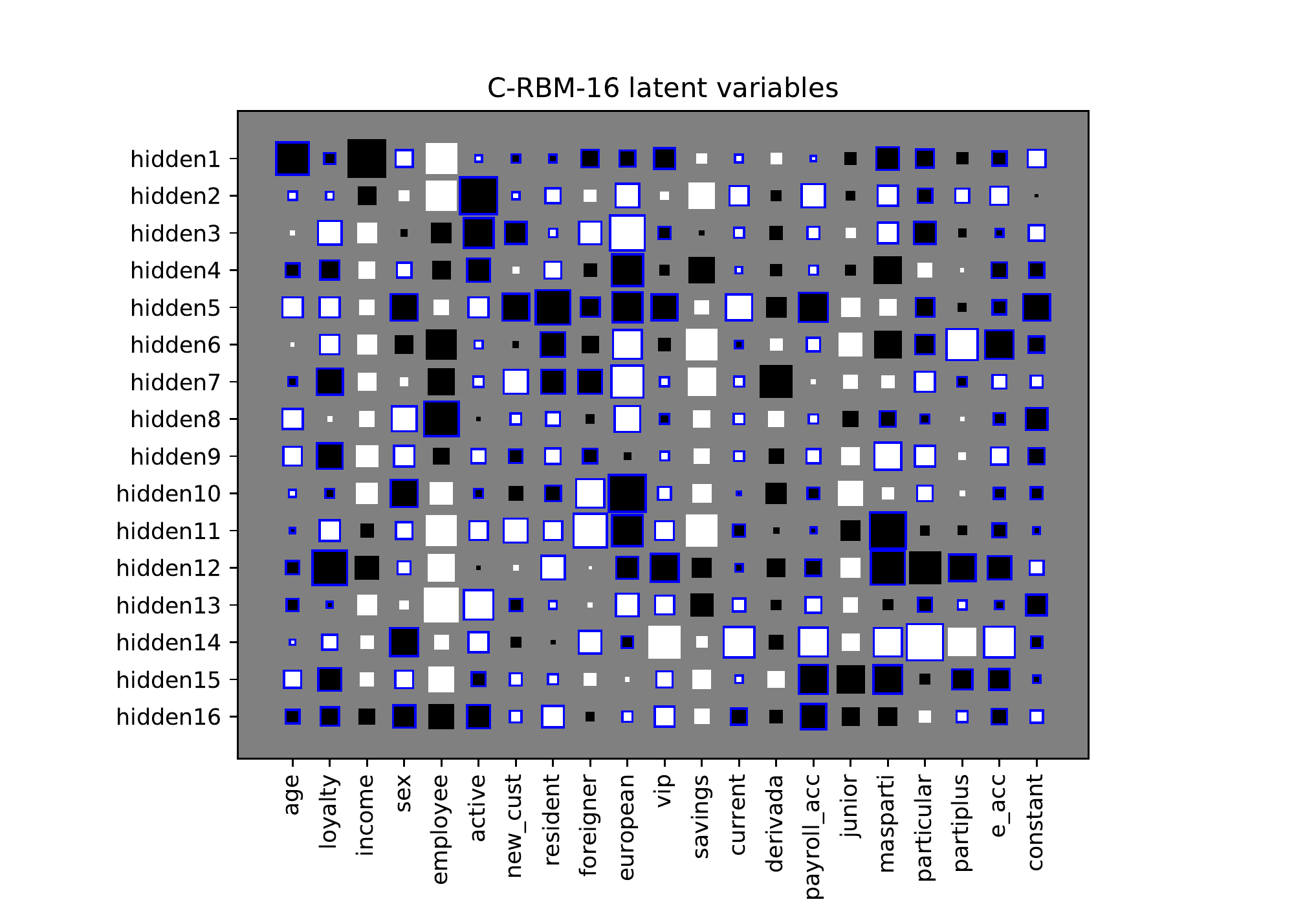}
			\caption{}
			\label{matrix_08}
		\end{subfigure}
	}
	\caption{(a) C-RBM model with 16 latent variables. (b) Latent variable relationship parameters. White: +ve values, Black: -ve values, Blue: >95\% significant}
	\label{fig_crbm8}
\end{figure}

To evaluate the efficiency of the models, we used a Hinton diagram \citep{Bremner1994} to analyze the parameter strengths between independent and dependent variables. We plot the parameter values and significance with choice on the y-axis and independent variables on the x-axis \citep{Bremner1994}. A Hinton diagram is often used in model analysis where the dimensionality of the model is high and provides a simple visual way of analyzing each vector. Figs. \ref{matrix_00} through \ref{fig_crbm8} shows the parameter estimates of the completed training stage of the different models. The Hinton matrix shows the influence of each independent variable on each alternative or latent variable. Statistically significant (>95\% confidence bound) parameters are highlighted in blue. The values along the x-axis are normalized with zero mean and unit variance. The 13 financial product choices are listed on the y-axis. The estimated parameters and bias of the C-RBM prediction model $\textbf{B}$, $\textbf{D}$ and $\bm{c}$ are projected onto the Hinton diagram (Figs. \ref{matrix_01}, \ref{matrix_03}, \ref{matrix_05} and \ref{matrix_07}) while parameters $\textbf{A}$ and $\bm{d}$ representing the parameters and bias for the latent variable with respect to the alternatives shown in Figs. \ref{matrix_02}, \ref{matrix_04}, \ref{matrix_06} and \ref{matrix_08}.  $\bm{c}$ and $\bm{d}$ are the constants with respect to the observed and hidden layer respectively. The signs and value of each parameter corresponds to the size and colour of the patches in the matrix, with white and black representing positive and negative signs respectively. Statistical significance (t-test) of each parameter is calculated using $\frac{\theta}{\sqrt{\sigma}}$, where $\sigma$ is the inverse of the Hessian of the log likelihood with sample size adjustment with respect to the parameters.

\begin{table}[t]
	\centering
	\caption{Model training results}\label{table_modelstats}
	\begin{tabular}{l l l l l l l}
		\toprule
		Model& latent variables & Validation error & log-likelihood & $\rho^2$ & no. of params & BIC\\
		\midrule
		MNL          &(baseline) & 0.4454 & -206808 & 0.546 & 273 & 416915\\
		CRBM &$J=2$  & 0.4360 & -203558 & 0.553 & 341 & 411237\\
		&$J=4$   & 0.4338 & -202066 & 0.556 & 409 & 409075\\
		&$J=8$   & 0.4323 & -200846 & 0.559 & 545 & 408279\\
		&$J=16$  & 0.4318 & -200223 & 0.560 & 817 & 410321\\
		\bottomrule
	\end{tabular}
\end{table}

\begin{table}[t]
	\centering
	\caption{Parameter sensitivity rank and standard error difference for estimated parameters $\textbf{B}$ for sampling-based sensitivity analysis} 
	\label{sensitivity}
	\resizebox{\textwidth}{!}{
		\begin{tabular}{l c c l c c l c c l c c l}
			\toprule 
			& \multicolumn{3}{c}{RBM 2 LV} & \multicolumn{3}{c}{C-RBM 4 LV} & \multicolumn{3}{c}{C-RBM 8 LV} & \multicolumn{3}{c}{C-RBM 16 LV}\\ 
			sample size & $n$ & $n_s$ &&  $n$ & $n_s$ &&  $n$ & $n_s$&& $n$ & $n_s$\\
			&&  &std. err. &&&std. err.  && &std. err.  &&  &std. err.  \\
			parameter & \multicolumn{2}{c}{rank} &diff.& \multicolumn{2}{c}{rank} &diff.&\multicolumn{2}{c}{rank} &diff.&\multicolumn{2}{c}{rank} &diff.\\
			\midrule 
			$\beta$age&15&15&49.30&15&12&0.52&11&11&0.99&11&12&0.64\\ 
			$\beta$loyalty&18&14&59.36&14&15&0.38&15&15&0.82&15&17&0.48\\ 
			$\beta$income&\cellcolor[gray]{0.65}3&\cellcolor[gray]{0.65}3&3712.99&\cellcolor[gray]{0.65}3&\cellcolor[gray]{0.65}3&26.67&\cellcolor[gray]{0.65}3&\cellcolor[gray]{0.65}3&43.00&\cellcolor[gray]{0.65}3&\cellcolor[gray]{0.6}2&35.82\\ 
			$\beta$sex&12&13&67.51&13&14&0.41&14&13&0.91&14&15&0.52\\ 
			$\beta$employee&\cellcolor[gray]{0.75}5&\cellcolor[gray]{0.6}2&4267.79&\cellcolor[gray]{0.6}2&\cellcolor[gray]{0.7}4&13.74&\cellcolor[gray]{0.75}5&\cellcolor[gray]{0.75}5&21.27&\cellcolor[gray]{0.7}4&\cellcolor[gray]{0.7}4&33.74\\ 
			$\beta$active&21&16&47.92&16&19&0.20&19&19&0.34&19&19&0.26\\ 
			$\beta$new\_cust&\cellcolor[gray]{0.8}6&12&53.93&12&\cellcolor[gray]{0.85}7&1.49&\cellcolor[gray]{0.9}8&\cellcolor[gray]{0.9}8&1.34&\cellcolor[gray]{0.9}8&\cellcolor[gray]{0.95}9&0.91\\ 
			$\beta$resident&16&20&16.61&20&20&0.19&20&20&0.31&20&20&0.23\\ 
			$\beta$foreigner&\cellcolor[gray]{0.9}8&17&29.15&17&\cellcolor[gray]{0.9}8&1.43&\cellcolor[gray]{0.95}9&10&0.76&\cellcolor[gray]{0.85}7&\cellcolor[gray]{0.85}7&1.35\\ 
			$\beta$european&17&20&16.62&20&20&0.19&21&20&0.31&21&20&0.23\\ 
			$\beta$vip&20&10&122.66&10&16&0.33&16&12&0.99&16&13&0.68\\ 
			
			$\beta$savings&\cellcolor[gray]{0.6}1&\cellcolor[gray]{0.6}1&34177.13&\cellcolor[gray]{0.6}1&\cellcolor[gray]{0.6}1&258.41&\cellcolor[gray]{0.6}2&\cellcolor[gray]{0.6}1&255.12&\cellcolor[gray]{0.6}2&\cellcolor[gray]{0.6}1&181.81\\ 
			$\beta$current&\cellcolor[gray]{0.85}7&11&64.19&11&13&0.41&12&18&0.38&12&16&0.39\\ 
			$\beta$derivada&\cellcolor[gray]{0.7}4&\cellcolor[gray]{0.7}4&3112.38&\cellcolor[gray]{0.7}4&\cellcolor[gray]{0.75}5&4.70&\cellcolor[gray]{0.7}4&\cellcolor[gray]{0.7}4&19.67&\cellcolor[gray]{0.75}5&\cellcolor[gray]{0.75}5&2.82\\ 
			$\beta$payroll\_acc&\cellcolor[gray]{0.95}9&18&24.91&18&18&0.29&18&17&0.52&18&18&0.41\\ 
			$\beta$junior&\cellcolor[gray]{0.6}2&\cellcolor[gray]{0.75}5&1759.26&\cellcolor[gray]{0.75}5&\cellcolor[gray]{0.6}2&58.29&1&\cellcolor[gray]{0.6}2&45.32&\cellcolor[gray]{0.6}1&\cellcolor[gray]{0.65}3&22.43\\ 
			$\beta$masparti&11&\cellcolor[gray]{0.85}7&185.94&\cellcolor[gray]{0.85}7&\cellcolor[gray]{0.95}9&1.41&\cellcolor[gray]{0.85}7&\cellcolor[gray]{0.8}6&4.99&\cellcolor[gray]{0.95}9&\cellcolor[gray]{0.8}6&2.29\\ 
			$\beta$particular&14&\cellcolor[gray]{0.9}8&166.56&\cellcolor[gray]{0.9}8&11&0.61&13&14&0.83&13&14&0.53\\ 
			$\beta$partiplus&10&\cellcolor[gray]{0.8}6&189.75&\cellcolor[gray]{0.8}6&10&0.65&10&\cellcolor[gray]{0.95}9&1.51&10&10&0.86\\ 
			$\beta$e\_acc&19&\cellcolor[gray]{0.95}9&159.38&\cellcolor[gray]{0.95}9&17&0.33&17&16&0.82&17&11&0.91\\ 
			bias&13&19&19.07&19&\cellcolor[gray]{0.8}6&3.17&\cellcolor[gray]{0.8}6&\cellcolor[gray]{0.85}7&3.35&\cellcolor[gray]{0.8}6&\cellcolor[gray]{0.9}8&0.48\\ 
			\bottomrule
		\end{tabular}
	}
\end{table}

\section{Analysis}
\label{S:6}
\subsection{Characteristics of latent variables}
We can characterize each hidden unit with the explained significance and strengths represented by the weights $\textbf{D}^\top$. $\textbf{D}^\top$ is a parameter vector that indicates the linear contribution of each latent variable and a constant $\bm{d}$, such that each alternative can be described as a utility function of latent variables: $y = \textbf{D}h + \bm{d}$. 

\begin{figure}
	\makebox[\linewidth][c]{
		\begin{subfigure}{0.6\textwidth}
			\includegraphics[width=1\textwidth]{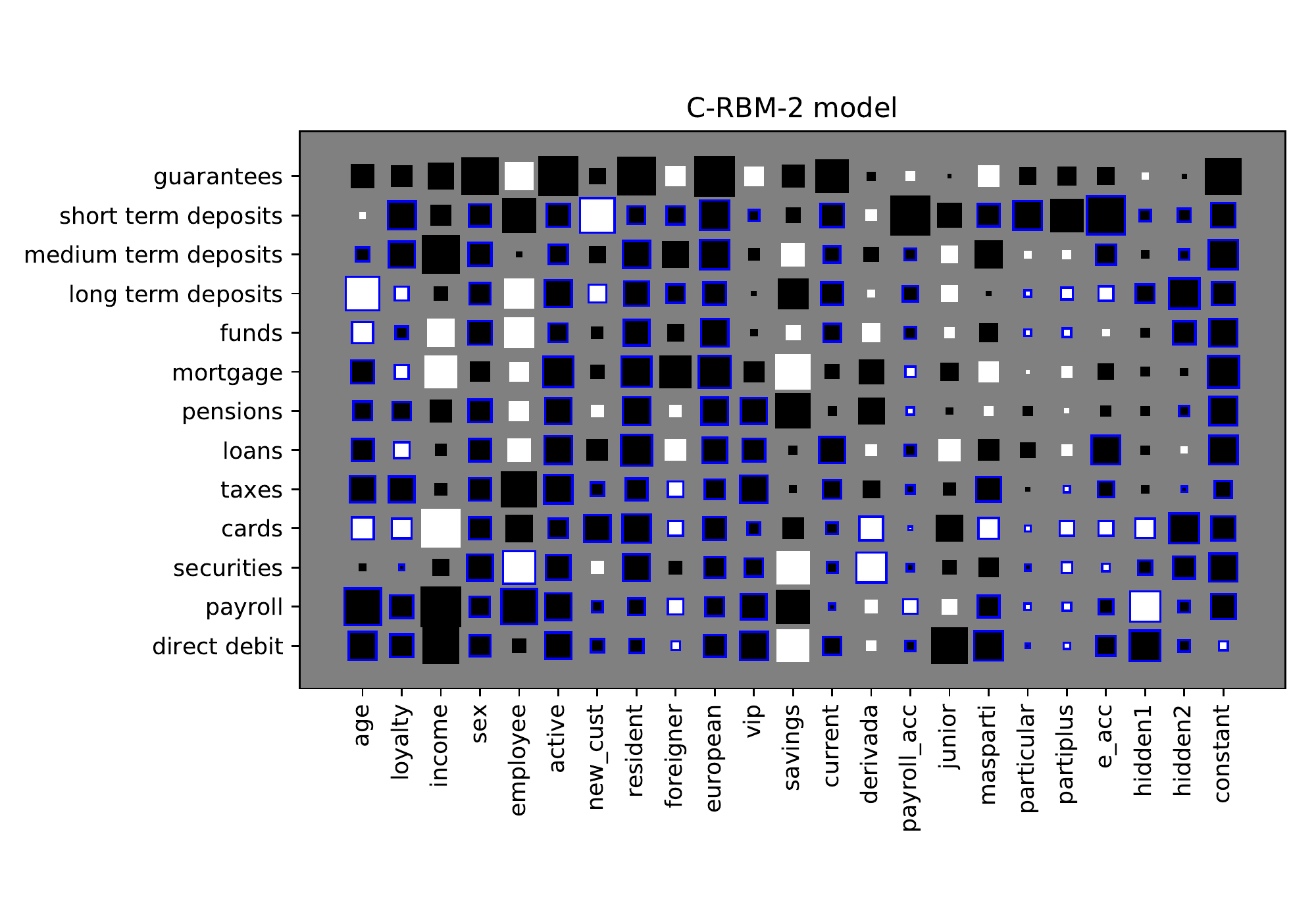}
			\caption{}
			\label{matrix_01}
		\end{subfigure}
		\begin{subfigure}{0.6\textwidth}
			\includegraphics[width=1\textwidth]{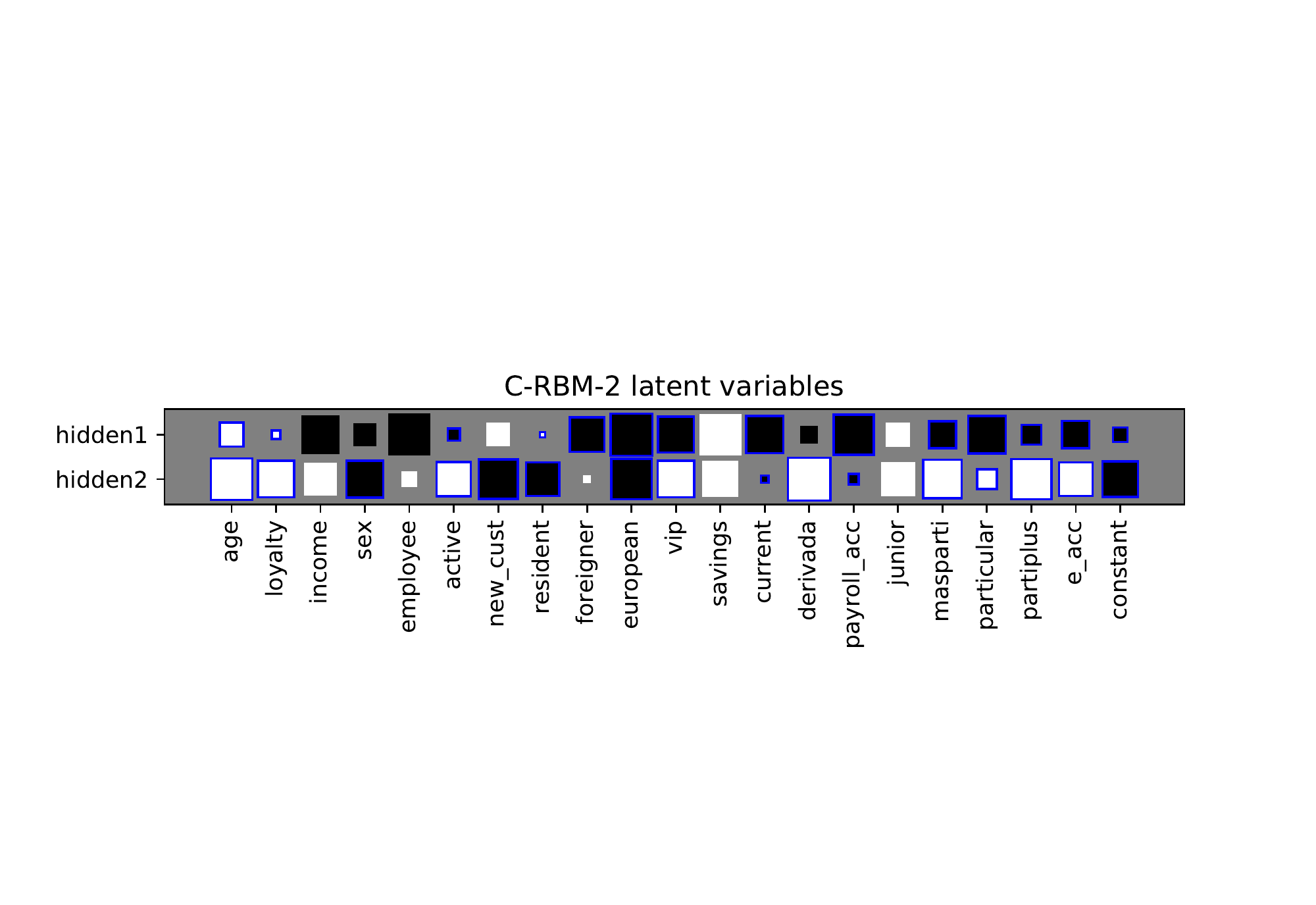}
			\caption{}
			\label{matrix_02}
		\end{subfigure}
	}
	\caption{(a) C-RBM model with 2 latent variables. (b) Latent variable relationship parameters. White: +ve values, Black: -ve values, Blue: >95\% significant}
	\label{fig_crbm2}
\end{figure}

For example, C-RBM-2 latent variable \textit{hidden1} is characterized by individuals who are of working age, non-EU foreign citizens with non-VIP status and does not own any special accounts. We can therefore infer this latent variable that indicates a `savings driven attitude' (see Fig \ref{matrix_02}).  From the model results, population with such characteristics have a positive preference of purchasing a payroll product and a low motivation of purchasing a (credit/debit) card product as indicated in Fig \ref{matrix_02}. Likewise in latent variable \textit{hidden2}, it is represented by older, loyal customers who are VIP and have held various account types over their lifetime. This latent variable can be inferred to as `self-reliance attitude' and are indication of the population who are less likely to purchase long term deposits, funds, securities and card products. The C-RBM with latent variables outperforms the MNL model, however, the performance increase from increasing the number of latent variables past 4 LV, is small. This would suggest that the upper bound of latent representative capacity is reached with just a small number of latent variables. Using 2 or 4 latent variables would be sufficient for significant improvement over a MNL structure.

From the presented results, it is clear that the C-RBM models differ significantly from the MNL model in terms of parameters which are strong and significant. This result seems to be broad-based in the sense that it is not dictated by the number of hidden units and signifies that the observed distribution has some latent factors that can be explored. However, we should mention that the training parameter initialization may have a small random effect on the model. Note that in the parameter plots, the signs and strength contribution to the choice model differ from model to model which may indicate that model training may be stuck at a local optima. This also suggest that the hidden and observed layer have different scale \citep{glorot2010understanding}. What is suggested in \citet{he2015delving} is to increase learning rate to improve convergence, but that would result in overgeneralization and loss of expressive power in the hidden units. We posit a middle-of-the-road solution should have adequate model accuracy and generalization over a large population.

We performed 2-fold cross-validation analysis and determined that the residual from model fit is not significant, therefore the model is robust to changes in input data -- this is further confirmed with a sensitivity analysis presented in the following section. In the parameter plots, we can see the values and signs correspond to the strength of each variable. For instance, the parameters for \textit{Guarantees} choice are not significant, since the distribution is very low (0.002\%). The latent models show similar results. For C-RBM with 2 and 4 hidden units, almost all of the parameters are significant, except for \textit{income}, \textit{employee}, \textit{savings}, \textit{derivada} and \textit{junior} variables. This can be attributed to the small mean values (and high deviation).

\subsection{Sensitivity of parameter estimates}
The versatility and effectiveness of parameter estimates are determined by a sensitivity analysis of the model output. Methods of sensitivity analysis include variance based estimator, sampling based and differential analysis \citep{helton1991sensitivity,saltelli2000sensitivity,saltelli2010variance}.``Sensitive'' parameters are those whose uncertainty contributes substantially to the test results \citep{helton1991sensitivity,hamby1994review}. The model is sensitive to input parameters in the variability associated with the input variable resulting in a large output variability. Sensitivity ranking sorts the input parameters by the amount of influence it has on the model output and the disagreement between rankings measures the parameter sensitivity to changes to the input \citep{hamby1994review}. 

We first define a list of parameters used in the model by their standard errors calculated over the full dataset. In large dataset sensitivity analysis, a key concern is the computational cost needed to complete the analysis, hence we use a sampling based approach as a cheap estimator to the output \% difference of the parameter minimum and maximum value. Random sampling (e.g. simple random sample, Monte Carlo, etc.) generates distributions of input and output to assess model uncertainties \citep{helton1991sensitivity}. Analyzing the sampling effects can provide information of the overal model performance since parameter sensitivity depends on all parameters which the model is sensitive and therefore the importance of each parameter \citep{hamby1994review}. 

Consider that the C-RBM model is represented by $y=f(x,h)$, where $x$ and $h$ are the input vectors of observed and latent variables respectively and $y$ is the model output. We suppose that the model $f(\cdot)$ is a complex, highly non-linear function such that we cannot completely define the way the C-RBM model responds to changes in input variables. Also, $h$ is dependent on $x$ through a submodel previously shown in Fig. \ref{icmc_fig01}. The analysis involves independently and randomly generated sample with size $n_S = 0.1n$ (10\% random sample draw), where $n=76,141$ is the total number of observations. The model performance is considered by sampling stability of variable parameters.  Sensitivities are also assessed for size of hidden units used in generating the C-RBM models and indicates what number of latent variables (hyperparameter) is required for model identifiability. Since the model was applied using a multinomial logit approach instead of a conditional logit, this resulted in a very large number of parameters. Thus the effect of relative changes to the number of distributed parameters gives the range of variance across each explanatory variable and number of hidden units used. Table \ref{sensitivity} shows the effects of sampling on the sensitivity and stability of the model observed parameters on the theoretical values and size of latent variables. Notice that the relative difference in standard error between the full and sampled model decreases when number of latent variables increases. This shows that the C-RBM model with high synthetic latent variables are robust to changes to input values through sampling. Additionally, the parameter sensitivity rank across variables also becomes more consistent and therefore,we show that RBM models are efficient in obtaining good latent variables with low generalization error. The significant decrease in standard error difference from 2 LV to 4 LV may indicate that the number of latent variables used in the models has a lower bound on the generalization error, which implies that we need careful consideration on $h$ for obtaining efficient but yet accurate exact values of $\beta$ without losing model interpretability.

\section{Conclusion}
\label{S:7}
This study analyzes alternative means of latent behaviour modelling in the absence of attitudinal indicators. In ICLV models, specialized surveys have to be constructed with attitudinal questions to model latent effects on the decisions. While it has been one of the more popular method in discrete choice analysis, there are several disadvantages to it. First, attitudinal questions are subjective and the behaviours are subjected to changes over time. Next, existing datasets that have no attitudinal questions cannot leverage on the ICLV model, thus latent effects cannot be utilized. We explore generative modelling of the choice distribution to uncover latent variables using machine learning methods, without measurement indicators. We hypothesized that latent effects can be obtained not only from attitudinal questions, but also from the posterior choice distribution. In effect, we are modelling latent components that fits the real choice distribution rather than achieving good statistics on subjective models. For example, there could possibly be some mean behaviour that dictates a more probable influence on purchases given some latent variables.

For this method to be effective, certain conditions have to be present: First, difficultly to get a good discriminative prediction result using only the provided explanatory variables. In this scenario, the C-RBM models were able to learn good latent variable representation and improve the model fit and prediction accuracy while providing latent variable inferrability. Next, when the data lacks attitudinal survey data, this method can find latent effects without the use of subjective measurement indicators.  

The current limitations of this study are the absence of choice dynamics or explanatory variable dynamics, i.e. changes over time or multiple choices for the same individual was not considered, but can be brought in. The underlying RBM is capable of dynamics. We hypothesize that this may improve the model significantly, but we are still looking for ways to incorporate dynamics into our C-RBM model. In recent studies, we have seen dynamic frameworks such as recurrent neural networks used in modelling temporal data \citep{taylor2007modeling,mnih2012conditional}. Finally, it is worth noting that as the number of latent variable increases, the number of estimated parameters increases exponentially. This will pose problems in large datasets and the ability to reduce dimensionality will give a significant benefit to efficient use of model parameters. In our observation, performing cross-validation or model selection with lowest validation error is a justifiable method to prevent overfitting using all the parameters. In the future, we would also look at the possibility of introducing deep learning architecture to choice modelling by stacking RBMs \citep{otsuka2016deep}.  

While ICLV model are optimized to predict the effects of latent constructs on the choice model using measurement indicators to guide latent parameters selection, this method uses observed decisions as an influence source for optimizing latent variables through machine learning. This is not to say that we do not agree with using measurement indicators which may often be subjective and may raise mis-specification problems and when explanatory variables are poor predictors, ICLV models can improve latent effects on choice models \citep{vij2016and}. However, latent effects may not only be present in attitudes and perceptions, but also in the direct observation of choices. Our current work explores the use of posterior choice distribution for latent behaviour modelling. Generative modelling in DCA is inspired by state-of-the-art machine learning algorithms that performs unsupervised feature extraction from unlabelled data used in classification problems \citep{hinton1984boltzmann}. In circumstances when attitudinal variables are not available, we have a strong reason to believe that the generation of latent factors are important and effective in building a discrete choice model. 

A future study that would be of interest is to extend this method to datasets with attitudinal questions and survey. For example, inter-city rail survey \citep{sobhani2017innovative}, and perform an analysis on both RBM and ICLV methods to obtain the generalization error of attitudinal survey models. A comparative study would provide a foundation for analysis of various latent behaviour models through graphical and algorithmic methods and provide guidance not only in selecting the appropriate latent variables, but also direct research effect to more promising directions.

\section*{Acknowledgements}
This research is in part funded by Ryerson University PhD Fellowship and by Fonds de recherche du Qu{\'e}bec - Nature et technologies (FRQ-NT) with team grant No. 2016-PR-189250

\clearpage




\bibliographystyle{model1-num-names}
\bibliography{sample.bib}







\end{document}